%% file: ms.tex
\documentclass[12pt, oneside]{article} 
\usepackage{amscd}
\usepackage{amsmath}
\usepackage{amssymb}
\usepackage{amsthm}
\usepackage{verbatim}
\usepackage[utf8]{inputenc}
\usepackage{geometry}     
\usepackage{subfig}
\usepackage[pdftex]{graphicx}				
\usepackage{setspace}
\usepackage{physics}
\usepackage[numbers]{natbib}
\usepackage{pdfpages}
\usepackage{bm}
\usepackage{wrapfig}				
\usepackage{afterpage}

\usepackage{bibentry} 
\usepackage{hyperref}

\graphicspath{{images/}{{\subfix{images/}}}}
\usepackage{tikz}
\usepackage{tikz-3dplot}
\usetikzlibrary{shapes,calc,positioning}
\tdplotsetmaincoords{70}{120}
\usepackage{siunitx}
\usepackage{adjustbox}
\usepackage{xcolor} 
\usepackage{blindtext}
\usepackage{subfiles} 
\usepackage{multicol,multirow}  
\usepackage{mathtools, nccmath}

\usepackage{lipsum}			
\usepackage{multicol}
\usepackage{graphicx,float}
\usepackage{xr}

\makeatletter
\newcommand*{\addFileDependency}[1]{
  \typeout{(#1)}
  \@addtofilelist{#1}
  \IfFileExists{#1}{}{\typeout{No file #1.}}
}
\makeatother

\DeclareUnicodeCharacter{88}{\-i}
\begin{document}

\begin{titlepage}
\begin{center}

\vspace*{2cm}

{\Large \textbf{Romantic-Computing}} 

\vspace{5cm}
{Honors Thesis\\ by}

\vspace{2cm}
{\large \textbf{Elizabeth Horishny}\\ Faculty Advisor: Dr. Simona Doboli}
\vfill

April 2022


\end{center}
\end{titlepage}



\section{Abstract}
\input{Chapters/Abstract}


\section{Introduction}
\input{Chapters/Introduction}

\section{Related Work}
\input{Chapters/RelatedWork}

\section{Data Description}
\input{Chapters/Data}

\section{Methods}
\input{Chapters/Methods}

\section{Results}
\input{Chapters/Results}

\section{Conclusions}
\input{Chapters/Conclusions}

\appendix 
\section{Appendix with Additional Poems, Tables, and Charts}
\input{Chapters/Appendix}

\newpage
\bibliographystyle{unsrtnat}

\bibliography{bibliography}
\end{document}

%% file: Chapters/Abstract.tex
In this paper we compare various text generation models' ability to write poetry in the style of early English Romanticism. These models include: Character-Level Recurrent Neural Networks with Long Short-Term Memory, Hugging Face's GPT-2, OpenAI's GPT-3, and EleutherAI's GPT-NEO. Quality was measured based syllable count and coherence with the automatic evaluation metric GRUEN. Character-Level Recurrent Neural Networks performed far worse compared to transformer models. And, as parameter-size increased, the quality of transformer models' poems improved. These models are typically not compared in a creative context, and we are happy to contribute.

%% file: Chapters/Introduction.tex
In this report, we will briefly introduce language models, and introduce some machine learning concepts. This will be followed by related work in the field. Consequently, we will discuss our data set in detail. Then, we will explain our implementations across each model. Afterwards, we will showcase poetry and make generalizations about how each model performed. Finally, we will state the conclusions made based on an analysis of our results. An appendix with additional poems, charts, and tables will be included at the end.\\
\subsection{Background}
Recurrent neural networks and transformer models are compared in the domains of translation, question-answering, and reading comprehension.\cite{huggingface} However, less work has been done analyzing the creative, or artistic, abilities of text generation models. This report compares several neural network implementations of language models. Namely, character-level recurrent neural networks with Long-Short Term Memory and pre-trained Transformer models. \\
By definition, a \textbf{language model} is a statistical probability distribution over a set of words. Given a sequence of tokens $w$ of length $m$, a language model assigns a probability  $ P(w_{1},...,w_m)$ to each token. At time-step $i$, a token is chosen based on the language model's probability distribution $\prod_{i=1}^{m-1}P(w_{i}|w_1,...,w_{i-1}) $. 
In contrast with conventional neural networks, recurrent neural networks (RNN) contain cycles within the network which allow reference of previous outputs. In other words, RNN allow neural networks to remember. Unfortunately, on its own, a RNN's memory, or context, becomes over-saturated with information. Over time, output recycled back into the hidden layer will either explode or become hidden. This problem is commonly referred to as the 'vanishing gradient' problem. \\ 
Long short-term memory, or LSTM, is one of many 'attention-mechanisms' built to prevent this problem. During each time-step, a hidden ‘short-term’ state is composed, retaining information which could have been lost. Essentially, at each time-step or token (think each word in a sentence) an LSTM unit will compute what should be remembered, and consequently, what should be forgotten. This output is sent back as input, alongside the subsequent input token, during the next time-step. \\
Unfortunately,  recurrent neural network architectures are still privy to the vanishing gradient problem. Especially when tasked with long texts. Furthermore, because recurrent neural networks process information token-by-token, they cannot process text concurrently. This makes RNNs extremely slow to train.
On the other hand, transformers, an alternative approach to generating text, analyze all tokens simultaneously. More accurately, each token is processed through an alignment score which determines what information is relevant in a given moment. Alignment scores are determined in a few primary ways. The most popular methods are additive or concat\cite{concat_transformers}, location-based, general and dot-product\cite{dot_transformers}. 
\begin{figure}[H]
\centering
\begin{minipage}{.5\textwidth}
  \centering
   \includegraphics[width=.77\textwidth]{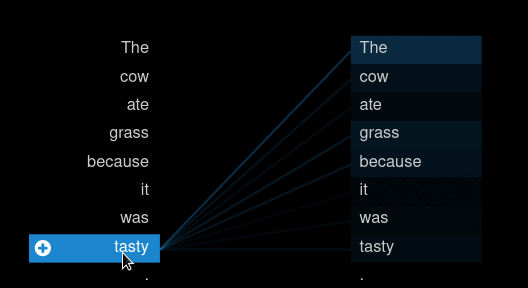}
  \label{fig:tasty}
\end{minipage}%
\begin{minipage}{.5\textwidth}
  \centering
   \includegraphics[width=.77\textwidth]{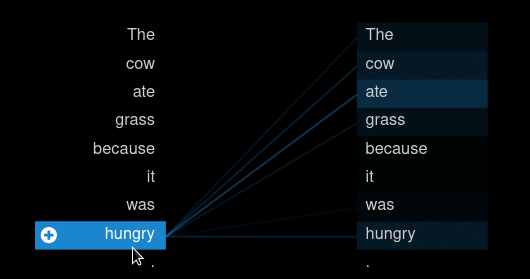}
    \label{fig:hungry}
\end{minipage}
\caption{Attention-mechanism visualization. Transformers differentiate the pronoun 'it' based on context.}
\end{figure} 
The era of Early English Romanticism was chosen because of it's multitude of quantifiable characteristics. Firstly, most poems follow a quatrain rhyme scheme. In other words, every four lines end in one of the following formats: AABB, ABAB, or ABBA. Additionally, rhyming line pairs will match in meter. In other words, when two lines share meter when the lines have the same number of syllables. Meter also encompasses the arrangement of stressed and unstressed syllables. But, this report only considers syllable count. 
Furthermore, many reoccurring themes are expressed within these poems. For instance, themes of nature, love, and night-time are abundant. With similar themes, we expect a specialized vocabulary capable of encapsulating Romanticism-era imagery.
Thereinby, poem-quality across all models will be based on general coherence in addition to conforming to this poetic structure. \\

%% file: Chapters/RelatedWork.tex
There have been many creative approaches to generating poetry using recurrent neural networks. \\
Deep-speare is a word-level recurrent neural network composed of 3 models, all of which have a unique purpose: a Language Model, a Pentameter Model, and a Rhyme Model. The Language Model builds on a LSTM encoder-decoder Model. Here, each word is encoded with orthographic information. The Pentameter Model is of an encoder-decoder architecture with attention as well. Using output from the Lanugage Model, the Pentameter Model suggests 10 candidate lines following accurate iambic pentameter. Finally, the Rhyme Model is trained to distinguish rhyming word pairs. In short, researchers found that their network was very good at grasping meter and rhyme. The surveyed layman population could not distinguish human written poems from generated ones. However, experts easily distinguished human poems from nonhuman ones. The network's focus on form and rhyme hurt its readability and emotion. To quote the paper, "real poets will break rules of form to create other effects."\cite{deepspeare} \\
In \textit{Generate and Revise}\cite{gen_rev}, researchers implemented a Reinforcement Learning approach to text generation. The intention of Generate and Revise is to generate poetry in such a way that mimics a human's revision process. Like Deep-speare, this system is composed of three neural network models- the Generator, Detector, and Prompter. Firstly, the Generator Model will compose a quatrain. Then, the Detector Model will seek out words that should be revised in order to fit the encoded rhyme scheme. Finally, the Prompter Model will generator a replacement word. The poem is then cycled between the Detector and Prompter Models. Results largely satisfied the intended quatrain rhyme scheme. In terms of coherence, results were measured against automatic text evaluation metrics BLEU and ROUGE. \\
The Recurrent Neural Network Poem Generator (RNNPG) approach was designed by Zhang et al.\cite{chinese_rnn} Here, poems were designed to fit a Chinese 5-character quatrain style. Each line is meant to follow different guidelines in terms of meter. Therefore, each token was encoded with its line number. Finally, poems were analyzed by humans in addition to automatic evaluator BLEU. Human ratings judged fluency, coherence, meaning, poetic-ness, and rank. Overall, human poems always scored slightly better than RNNPG poems in BLEU and human ratings. 

%% file: Chapters/Data.tex



Lord Byron and Percy Bysshe Shelley are noted as preeminent poets in early English Romanticism. For our data set, we used the entire bibliography of poets Percy Bysshe Shelley and Lord Byron, as listed by poetry catalog website \textit{allpoetry.com}.\cite{poetry.com} Using python package Beautiful Soup\cite{soup}, all poems were scraped from \textit{allpoetry.com} and into a single text file. In total, 628 poems were scraped with a fairly even split with 334 belonging to Percy Bysshe Shelley and 294 belonging to Lord Byron. In total, this text file was comprised of 80,409 lines of poetry, 338,387 words, 46,926 unique words, and 5,356,670 bytes. All models were either fully-trained or fine-tuned on this data set. Technically, our character-level recurrent neural network models were the only properly trained models. Most other models belonged to pre-trained transformer models and were only fine-tuned, or not tuned at all.

%% file: Chapters/Methods.tex
\subsection{Character-level Recurrent Neural Network with LSTM}
The first models we trained were a character-level recurrent neural network with long short-term memory layers. We chose to generate at the character-level instead of the word-level in order to compromise with the relatively small data set. The model was built in Python using PyTorch nn. Across all models, the learning rate was 0.02, the chunk length was 200, and the batch size was 100. Otherwise hidden-layer size varied between 100-200 neurons, and 2 or 3 LSTM units were included. Finally, training was determined based on perceived improvement during the training process. This varied between 6,000-24,000 epochs. Code was largely taken from \textit{https://github.com/spro/char-rnn.pytorch}\cite{spro}. 
\subsection{GPT-2}
The first Transformer model GPT-2, or Generative Pre-trained Transformer was released in 2011 by OpenAI. OpenAI’s GPT-2 Model is an evolved version of their previous GPT model- primarily due to its increase in parameters (117 million to 1.5 billion). GPT-2 was also trained on a much larger data set, utilizing a subset of Common Crawl- a web scraping archive.\cite{attention} Although a massive innovation,  fine-tuning still required the acquisition of data sets. And, the use of a training set comes with the implications of faulty generations. \\
GPT-2 Experiments were conducted using Hugging Face’s transformers library which uses a byte-lever version of Byte Pair Encoding and a vocabulary size of 50,257. The model weights were initialized from Hugging Face's GPT-2 weights\cite{huggingface}. The batch-size and sequence-length was 64. Code was primarily based off of a Kaggle tutorial by user Michaelarman\cite{gpt_kaggle}.\\
For our first model, we used the Hugging Face transformer library  with PyTorch’s DataLoader Class, for forming training and testing data sets as well as batching data.  Fine-tuning was conducted using Hugging Face libraries for 10 epochs. After 10 epochs, the model began to over-fit (validation score decreased while perplexity increased). \\
Another GPT-2 model used the Hugging Face transformer library in tandem with Fast.ai’s DataLoader and Loader classes. Fast.ai’s DataLoader structures data, like PyTorch’s DataLoader. But, it also automatically feeds data into the model as batches, and creates validation sets for training and testing. Furthermore, Fast.ai’s Learner class was used for fine-tuning. The Learner class is supplied with Fast.ai’s custom loss functions. CrossEntropyLossFlat() is the loss function we used to train for a total of 48 epochs. After 48 epochs, perplexity stopped significantly decreasing. \\
To recapitulate, both models used Hugging Face's GPT-2, but were fine-tuned using different methods.

\subsection{GPT-3}
\begin{figure}[H]
    \centering
    \includegraphics[width=0.8\textwidth]{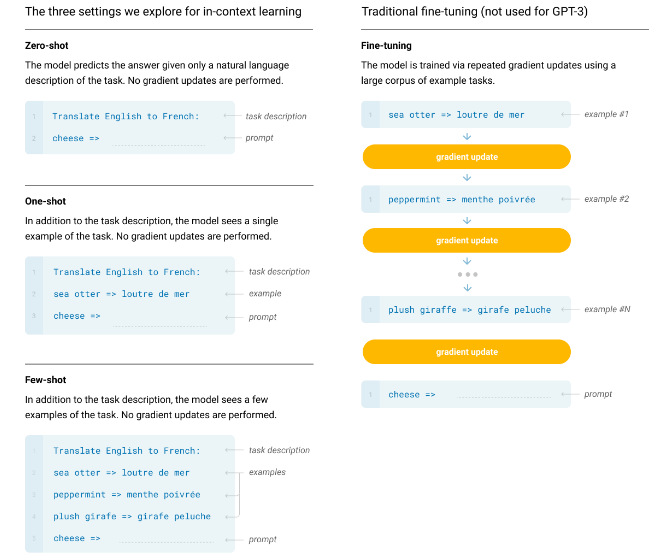}
    \caption{Examples of zero-shot, one-shot and few-shot, contrasted with traditional fine-tuning.\cite{gpt3}}
    \label{fig:shots_fired}
\end{figure}

OpenAI's GPT-3 uses the same foundation as GPT-2, but the number of parameters is increased from 1 billion to 175 billion. Essentially, GPT-3 is GPT-2, but scaled up by 10. See Table \ref{fig:transformer models} to see how they compare. \\ GPT-3’s training data primarily consisted of a larger section of the Common Crawl data set, which amounts to a trillion words. With this enormous dataset, GPT-3 completely eliminates the need for fine-tuning. As a matter of fact, GPT-3 gave rise to zero-shot learners. Whereas training or fine-tuning primes a model with examples, GPT-3 only needs a prompt. See Figure \ref{fig:shots_fired} for more examples of zero-shot prompts. And, although GPT-3 works with only a prompt, GPT-3 has been recorded to answer with higher accuracy given a few examples (few-shot learning).\cite{gpt3}

GPT-3 has yet to be fully released, but poems can be generated after applying for an API-key from \textit{https://beta.openai.com/}. This API-key was downloaded and used.

\subsection{GPT-NEO}
GPT-3 code has yet to be fully released to the public. In an attempt to supply an open-source few-shot learner, EluetherAI released Generative Pre-Trained Transformer NEO. It scales larger than GPT-2 in terms of parameter-size and context-size, but smaller than GPT-3.\cite{gptneo} Code was based off of an aitextgen tutorial.\cite{woolf-neo-code}
    
 \begin{table}[H]
     \begin{center}
        \begin{tabular}{llll}
        \hline
        \multicolumn{1}{|l|}{}       & \multicolumn{1}{l|}{GPT-2} & \multicolumn{1}{l|}{GPT-NEO} & \multicolumn{1}{l|}{GPT-3} \\ \hline
        \multicolumn{1}{|l|}{Vocabulary}   & \multicolumn{1}{l|}{50,257} & \multicolumn{1}{l|}{50,257} & \multicolumn{1}{l|}{50,257} \\ \hline
        \multicolumn{1}{|l|}{Layers} & \multicolumn{1}{l|}{48}    & \multicolumn{1}{l|}{n/a}      & \multicolumn{1}{l|}{96}    \\ \hline
        \multicolumn{1}{|l|}{Batch-size}   & \multicolumn{1}{l|}{512}    & \multicolumn{1}{l|}{n/a}    & \multicolumn{1}{l|}{3.2M}   \\ \hline
        \multicolumn{1}{|l|}{Context-size} & \multicolumn{1}{l|}{1024}   & \multicolumn{1}{l|}{2048}   & \multicolumn{1}{l|}{2048}   \\ \hline
        \multicolumn{1}{|l|}{Parameters}   & \multicolumn{1}{l|}{1.5B}   & \multicolumn{1}{l|}{2.7B}   & \multicolumn{1}{l|}{175B}   \\ \hline   
        \end{tabular}
         \caption{Comparison of Transformer models.}
         \label{fig:transformer models}
         \end{center}
    \end{table}
    
\subsection{GRUEN}
GRUEN, which loosely stands for \textbf{G}rammaticality, non-
\textbf{R}edundancy, foc\textbf{U}s, structure and coher\textbf{EN}ce, is an automatic evaluation metric for generated text. It measures grammatical correctness, the degree of focus across all sentences, whether each sentence flows from one to the next, and a text's redundancy or repetitiveness. GRUEN outperforms BLEU and ROUGE in terms of accessing accuracy.\cite{zhu2020gruen} We will be using GRUEN to check the legibility of our generated texts. \\
\subsection{Meter}
In order to quantify poetic meter, we wrote a Python script to count syllables in each line. Then, we calculated the standard deviation to see how close the syllables carried across a poem per line. In Byron and Shelley's poems, rhyming line pairs matched in syllable-length as well as rhyme. So, if a poem had the same number of syllables across all lines, the standard deviation would be zero, implying accurate use of meter. But, arose the reoccurring issues of \textit{'interrupted'} and '\textit{feeble beginnings'} generations. \textit{Interrupted generations} refer to poems which could have finished a line complying to the meter scheme, but were cut short by the max-length generation parameter. This case was determined when the final line was 4 or more syllables lacking from the previous line. In an \textit{interrupted generation}, the last line was not counted. Furthermore, a \textit{feeble beginning} occurred when the prompt was interpreted as the ending of a previous quatrain, rather than the beginning of a new one. In this case, if the first line was 4 or more syllables less than the previous line, it was discarded from the meter metric.
\subsection{Generation parameters}
For all models, generations were kept relatively short at a max-length of 300 characters or 75 words (length token differed per model). The intention was to produce a little more than a quatrain's worth of text. With at-most 300 characters, or 75 words, our results quantify how models compare in short-term generations. 
For Character-level RNN, GPT-2, and GPT-NEO generations, we settled on a temperature of 0.7 with the intention of producing creative, but not gibberish, text. For prompts, we used the starter sequences:  'Roses are red, ' and 'Dearest audience, '. The last two characters, a space and a comma, were essential to guarantee Char-RNN models wouldn't add additional characters to the prompt, resulting in a non-english word.
\\
GPT-3 was treated differently during generation because, rather than starter sequences, GPT-3 is a few-shot and zero-shot learner (Figure \ref{fig:shots_fired}. 
A generation parameter we could have tweaked was generation. Temperature controls how 'random' or expected a generation will become. The closer to 0 the more 'conservative' a generation will be. For a wider breadth of results, we used temperatures ranging from 0.7 to 1.5. Since GPT-3 accepts Zero-shot prompts rather than a beginner sequence, we queried the model with prompts such as: \textit{Write a poem in the style of Percy Bysshe Shelley. Start with 'Dearest audience, '}, \textit{Write a poem in the style of English romantic movement, Start with: 'Roses are red'.} All other parameters were default as supplied by OpenAI.
\\
\begin{figure}[H]
    \centering
    \includegraphics[width=.5\textwidth]{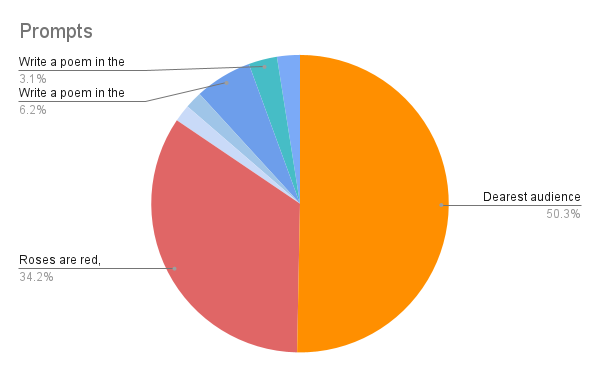}
    \caption{Distribution of Prompts for the experiment}
    \label{fig:prompts}
\end{figure}



%% file: Chapters/Results.tex

\newcommand{\rnnone}{        
\begin{figure}[H]
    \centering
    \textit{Dearest audience, mad shadows,}                     \\
\textit{Who lose like that gentle heart, and will die}      \\
\textit{That upon the burning vapoured in high,  }                   \\
\textit{Wrought out of his sea and are his und with power,} \\
\textit{And in the sea of sunless land to wear}             \\
\textit{The driver, at last displayed the winds.}           \\
\textit{The rapid that oneselv}     
    \caption{Char-RNN. 2 LSTM layers, hidden size 100.}
    \label{fig:rnn1}
\end{figure}
}

\newcommand{\hfroses}{        
\begin{figure}[h]
    \centering
\textit{Roses are red,}                     \\
            \textit{When the mountains are blue;}       \\
            \textit{When the stars are purple,}         \\
            \textit{And when the clouds are quivering,} \\
            \textit{And when the wind is wild;}         \\
            \textit{We feel that our bodies are a thing   }      \\
        \textit{    Immensely fixed on sensation,}               \\
           \textit{ And that, whatever we feel, }                \\
            \textit{Might be already.    }                      
    \caption{GPT-2 Pure Hugging Face model under prompt: 'Roses are Red, '.}
    \label{fig:hf_roses}
\end{figure}
}

\newcommand{\threepoem}{        
\begin{figure}[H]
    \centering
        \textit{Dearest audience,}            \\
        \textit{I welcome you all}            \\
        \textit{To my dark and dreary play}   \\
        \textit{Which I have written}         \\
        \textit{In hopes that it will}        \\
        \textit{Bring you some small measure} \\
        \textit{Of the joy that I have found} \\
        \textit{In its crafting.}             \\
        \textit{Please do not be alarmed}     \\
        \textit{By the blood and gore}        \\
        \textit{Which stains}                
   \caption{GPT-3 under prompt:\textit{ Write a poem in the style of Percy Bysshe Shelley. Start with: 'Dearest audience, '}}
    \label{fig:neo_shelley}
\end{figure}
}

\newcommand{\neopoem}{        
\begin{figure}[H]
    \centering
            \textit{Dearest audience!}                 \\
            \textit{She was a child of Nature;}        \\
            \textit{Her very bloom was of the air,}    \\
            \textit{And her very sparkles of the eye;} \\
            \textit{And she was the first of those,}   \\
            \textit{Who saw the sun rise}    
    \caption{GPT-NEO under prompt 'Dearest audience, '}
    \label{fig:neo_poem}
\end{figure}
}


\subsection{Character-Level Recurrent Neural Network}
Overall, there was no noticeable difference between RNN models, regardless of the number of LSTM layers, or the size of hidden layers. Therefore, all character-level RNN results will be considered analogously. Overall, generations mostly conformed to an English vocabulary. However, generations often lacked in semantic meaning. Largely, generations used flowery language to illicit an enigmatic, vaguely romantic, theme. Additionally, hints of a rhyme scheme would peek out in the occasional poem (in Figure \ref{fig:rnn1}- die/high, and perhaps wind if pronounced  \textit{wīnd}), but rhymes appeared more accidental than premeditated. See Figure \ref{table:meter-gruen-rnn} to see how all RNN models differed with regards to Meter and GRUEN. 
\rnnone
\subsection{GPT-2}
Originally, we set out to compare two different versions of Hugging Face's GPT-2 model, with minute differences in fine tuning. However, after collecting results, there was not a clear difference in quality between the pure Hugging Face model and the Hugging Face + Fast.ai hybrid. Furthermore, the jump in text quality is immense compared to the Character-Level Recurrent Neural Network. Average GRUEN went up by 200\%, and standard meter went down about 35\%. Again, rhyme schemes were inconsistently followed. But, instances of rhyming did occur (see Figure \ref{fig:hf_rhyme}). Refer to Table \ref{table:gpt-two-anal} to see how the two models compare between meter and GRUEN metrics.
\hfroses
\subsection{GPT-NEO}
GPT-NEO's generations were majorly improved in comparison to the GPT-2 models. Interestingly, this model performed the best on GRUEN. Meter was also marginally better than GPT-2. The most impressive quality of GPT-NEO, is within its long-term poems with a consistent theme from beginning to end. Although, this wasn't fully explored do to this experiments focus on short generations. See Figure \ref{fig:neo_long} for an example of a longer GPT-NEO poem. 
\neopoem
\subsection{GPT-3}
Finally, GPT-3 was the most impressive model. Its ability to understand zero-shot prompts can outperform a person tasked with the same problem. GPT-3 had the best understanding of meter, doing twice as good as Char-RNN. The only downside of GPT-3 are the bland, or conservative, generations. The choice in increasing temperature for this model was chosen in order to deter this, but "boring" generations still arose. See Figures \ref{fig:three_love} and \ref{fig:three_boring}) for a few examples of GPT-3's worst poems. \\
Additionally, with regard to performing worse than GPT-NEO on GRUEN: Although GPT-NEO had a higher score, from human analysis, poems generated by GPT-NEO were no more legible than those generated by GPT-3.  This is perhaps because GPT-3 followed a popular "Roses are red" poem structure that is recognizable by most people, but technically grammatically incorrect.
\threepoem

\begin{table}[H]
\centering
\begin{tabular}{|l|l|l|l|l|}
\hline
                                            & \textbf{CHAR-RNN} & \textbf{GPT-2} & \textbf{GPT-NEO} & \textbf{GPT-3} \\ \hline
\textbf{Mean GRUEN}                         & 0.18              & 0.36           & .48              & .39            \\ \hline
\textbf{Median GRUEN}                       & 0.18              & 0.39           & .53              & .39            \\ \hline
\textbf{Mean Meter $\sigma$}   & 3.06              & 2.39           & 2.18             & 1.23           \\ \hline
\textbf{Median Meter $\sigma$} & 2.85              & 2.08           & 1.76             & 1.15           \\ \hline
\end{tabular}
\caption{Averages of GRUEN and Meter across all models.}
\end{table}

%% file: Chapters/Conclusions.tex
Across all models, some semblance of poetic form was learned. Only GPT-2 would occasionally generate text formatted in a block of text (no new lines). Furthermore, all models picked up on the importance of matching lines syllabic-ally. This was less apparent with the character-level recurrent neural network models, however, more research is needed. Character-level RNN's were trained on far less data than any of the transformer models, since the transformer models were pre-trained. Perhaps if more training data was added, character-level RNN's would improve better in terms of meter. Same goes for GRUEN; character-level RNN's performed far poorer in GRUEN than any transformer models, but may have scaled better if their training vocabulary was increased.
\\
Additionally, very few training parameters were tweaked. And, although altered parameters did not result in a noticeable change, the breadth of combinations of a recurrent neural network's architecture is immense. The same thought holds for generation parameters. It's possible that a different number of layers, length of training, and temperature could have resulted in entirely different GRUEN and meter analytics. \\
To recapitulate, all models learned some poetic form from the Lord Byron and Percy Bysshe Shelley data set. The importance of meter, or syllable-count, was also encapsulated by all models. There was no significant presence of rhyme within the results, but all models showed a capability of making two line pairs rhyme. In terms of coherence, character-level recurrent neural networks performed the worst, with poems which rarely showcased meaning. On the other hand, all transformer models were capable of dictating a poem with consistent meaning from beginning to end. As parameter size increased, GRUEN and perceived meaning did as well. 

%% file: Chapters/Appendix.tex
\newcommand{\rnntwo}{        
\begin{figure}[H]
    \centering
\textit{Dearest audience,}                                     \\
\textit{The thin the stern doft howling burn,}                 \\
\scriptsize\textit{The love who some be trinkled wide  thee the stars }             \\
\scriptsize\textit{Comes a which when the delight thy prouctory stars}    \\
\scriptsize\textit{The has own since, for the gaze which talt that self?} \\
\textit{Have pines unteath spectation atennal day}             \\
    \caption{Char-RNN. 2 LSTM layers, hidden size 100.}
    \label{fig:my_label}
\end{figure}
}

\newcommand{\hfrhyme}{        
\begin{figure}[H]
    \centering
  \textit{Dearest Audience, to my longing sight}              \\
            \textit{Is more than thou canst be with me -}               \\
            \textit{Farewell! and though the dream now hath wing'd me,} \\
            \textit{I'll not sleep--and cannot leave to thee.}          \\
            \textit{My life is Lara--when}                        
    \caption{GPT-2 Hugging Face and FastAI hybrid model under prompt 'Dearest Audience, '.}
    \label{fig:hf_rhyme}
\end{figure}
}
\newcommand{\threelove}{        
\begin{figure}[H]
    \centering
            \textit{Dearest audience,}                       \\
            \textit{I love you all so much,}                 \\
            \textit{I wish I could take you with me}         \\
            \textit{On all my journeys.}                     \\
            \textit{You are the stars that light up my sky,} \\
            \textit{The sun that warms my soul,}             \\
            \textit{The moon that guides my way.}            \\
            \textit{You are everything to me,}              
    \caption{GPT-3 under prompt: \textit{Write a poem in the style of the English romantic movement. prompt: 'Dearest audience, '}}
    \label{fig:three_love}
\end{figure}
}
\newcommand{\threeboring}{        
\begin{figure}[H]
    \centering
\textit{Roses are red,}         \\
\textit{Violets are blue,}      \\
\textit{I love you,}            \\
\textit{And I always will too.}
   \caption{GPT-3 under prompt: \textit{Write a poem in the style of Percy Bysshe Shelley. Start with: 'Roses are red, '}}
    \label{fig:three_boring}
\end{figure}
}

\newcommand{\neolongest}{        
\begin{figure}[H]
    \centering
        \textit{I think, in the end, it is better to die,}   \\
        \textit{And live on the sea, and be what you are--}  \\
        \textit{But die, and be what you wish to be.}        \\
        \textit{It is a sad reality--but true--}             \\
        \textit{That death is what makes us human,}          \\
        \textit{And that we are the consequence.}            \\
       \textit{But in the end, life is better than death--} \\
        \textit{And that the consequence is life's loss.}    \\
        \textit{I do not pretend to say what's best,}        \\
        \textit{But merely that the world is hell.}          \\
        \textit{I say that life is, and ever was,}           \\
        \textit{The last of human perfection.}               \\
        \textit{But I would not pretend to say it--}         \\
        \textit{To soothe my penitents' fears,}              \\
        \textit{That life could not be happiness;}
        \caption{GPT-NEO under no prompt.}
    \label{fig:neo_long}
\end{figure}
}

\hfrhyme
\neolongest

\threelove
\threeboring

\begin{figure}[H]
\centering
\begin{minipage}{.5\textwidth}
  \centering
   \includegraphics[width=.9\textwidth]{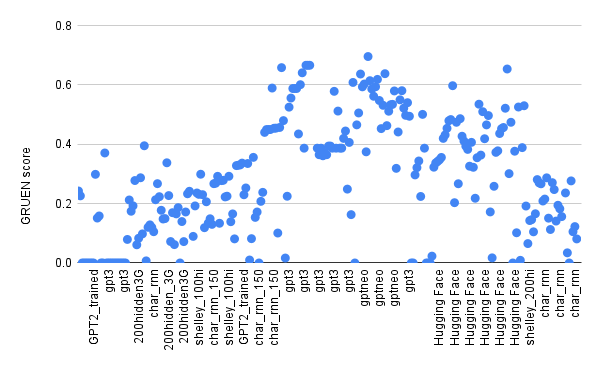}
\end{minipage}%
\begin{minipage}{.5\textwidth}
  \centering
   \includegraphics[width=.9\textwidth]{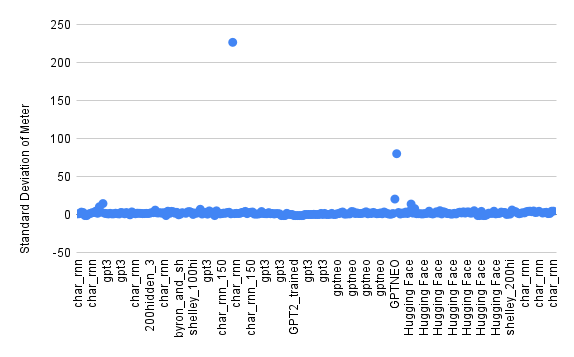}
\end{minipage}
\caption{Meter and GRUEN metrics across all RNN models. Little differentiation found between models.}
\label{table:meter-gruen-rnn}
\end{figure} 

\begin{table}[H]
\centering
\begin{tabular}{|l|l|l|l|}
\hline
                                   & GPT-2 & GPT-2 HF & GPT-2  HF+FAI \\ \hline
Mean GRUEN                         & 0.36  & 0.33     & 0.38          \\ \hline
Median GRUEN                       & 0.39  & 0.38     & 0.4           \\ \hline
Mean Meter $sigma$   & 2.39  & 2.4      & 2.37          \\ \hline
Median Meter $sigma$ & 2.08  & 2.18     & 2.05          \\ \hline
\end{tabular}
\caption{Little quantifiable difference between GPT-2 models.}
\label{table:gpt-two-anal}
\end{table}